%% file: root.tex
\documentclass[letterpaper, 10 pt, conference]{ieeeconf}
\IEEEoverridecommandlockouts 
\usepackage{hyperref}
\usepackage{threeparttable}
\usepackage{algorithm}
\usepackage{cite}
\usepackage{float}
\usepackage{caption}

\hypersetup{
    colorlinks=true,
    linkcolor=magenta,
    filecolor=magenta,      
    urlcolor=magenta,
}

\renewcommand{\thefigure}{\Roman{figure}}

\input{stachnisslab-latex}


\usepackage{bm}
\usepackage{multirow}
\usepackage{rotating}

\title{\LARGE \bf G2IA: Geometry-Guided Instance-Aware Retrieval and Refinement \\for Cross-Modal Place Recognition}

\author{{Xianyun Jiao$^{1 *}$, Jingyi Xu$^{1 *}$, Zhongmiao Yan$^{1}$, Xieyuanli Chen$^{2}$, and Ling Pei$^{1 \dag}$} 
  \thanks{$^{1}$Xianyun Jiao, Jingyi Xu, Zhongmiao Yan, and Ling Pei are with the Shanghai Jiao Tong University. $^{2}$Xieyuanli Chen is with the National University of Defense Technology. $^{*}$Equal contribution}
  \thanks{$^\dag$Corresponding author: ling.pei@sjtu.edu.cn}
}

\begin{document}
\maketitle

\IEEEpeerreviewmaketitle
\thispagestyle{empty}
\pagestyle{empty}

\begin{abstract}

Cross-modal place recognition (CMPR) enables camera-only robots to localize against pre-built LiDAR maps in autonomous navigation scenarios. This image-to-point-cloud setting is challenged by two coupled ambiguities: the modality gap between perspective RGB appearance and sparse metric geometry, and perceptual aliasing among urban places with similar roads, facades, intersections, and object arrangements. Instead of treating CMPR as a single global descriptor matching problem, we argue that reliable retrieval requires both geometry-aware representation alignment and fine-grained candidate verification. In this paper, we propose G2IA, a geometry-guided instance-aware framework for image-to-point-cloud place recognition. In the retrieval stage, visual geometry priors from VGGT and instance features are integrated to construct place descriptors that are more compatible with LiDAR-derived map representations. In the refinement stage, the retrieved candidates are re-ranked by explicitly verifying whether local instance shapes and their relative spatial layouts are consistent across modalities. Experiments on public benchmarks demonstrate that G2IA consistently improves image-to-point-cloud place recognition under different localization thresholds, and exhibits strong cross-dataset generalization. 

\end{abstract}

\section{Introduction}
\label{sec:intro}

Place recognition is a fundamental component of autonomous robot navigation~\cite{yin2022general, luo20243d, yin2024survey}. By recognizing whether a robot has revisited a known location, place recognition enables loop closing and global localization over long time spans~\cite{nie2024mixvpr++,uy2018pointnetvlad,ma2023cvtnet,zhou2023lcpr}. In many robotic and autonomous driving systems, LiDAR scans are used to build reference maps because point clouds provide accurate geometric structure and are relatively robust to illumination and appearance changes. However, during deployment, a robot platform may only have access to monocular camera images because of hardware cost, sensor failure, platform constraints, or sensor configuration changes. This practical mismatch motivates cross-modal place recognition (CMPR), where an RGB image is used as the online query and a LiDAR-based point-cloud map serves as the reference database.

Image-to-point-cloud place recognition is challenging because images and point clouds observe the same environment through fundamentally different signals. RGB images capture perspective appearance, color, texture, and partial fields of view, while LiDAR point clouds encode sparse metric geometry. This modality gap makes direct descriptor matching unreliable. Meanwhile, urban environments contain repeated roads, facades, intersections, and object arrangements, causing perceptual aliasing even when global scene descriptors appear similar. Therefore, reliable image-to-point-cloud place recognition requires more than learning a shared global descriptor. It must align cross-modal representations while also verifying whether the retrieved candidates preserve the local instance shapes and relative spatial layouts observed in the query.

Existing methods have made important progress by reducing the modality gap through depth-based conversion~\cite{lc2,I2p-rec}, bird's-eye-view (BEV) distillation~\cite{xu2025crossbev}, modality-invariant descriptor learning~\cite{liploc}, or foundation-model features~\cite{liploc,jiao2025inscmpr}. They perform retrieval primarily through global descriptor similarity. Although efficient, such global matching can confuse distinct places with similar holistic appearances, especially under strict localization requirements. Moreover, many methods treat modality conversion and feature extraction as separate stages, which limits their ability to exploit the geometric priors embedded in modern vision foundation models. These observations suggest that CMPR should be formulated as a retrieval-and-verification problem, in which geometry priors and instance-level consistency should guide cross-modal descriptor learning and candidate refinement.

\begin{figure}[t]
  \centering
  \includegraphics[width=1.00\linewidth]{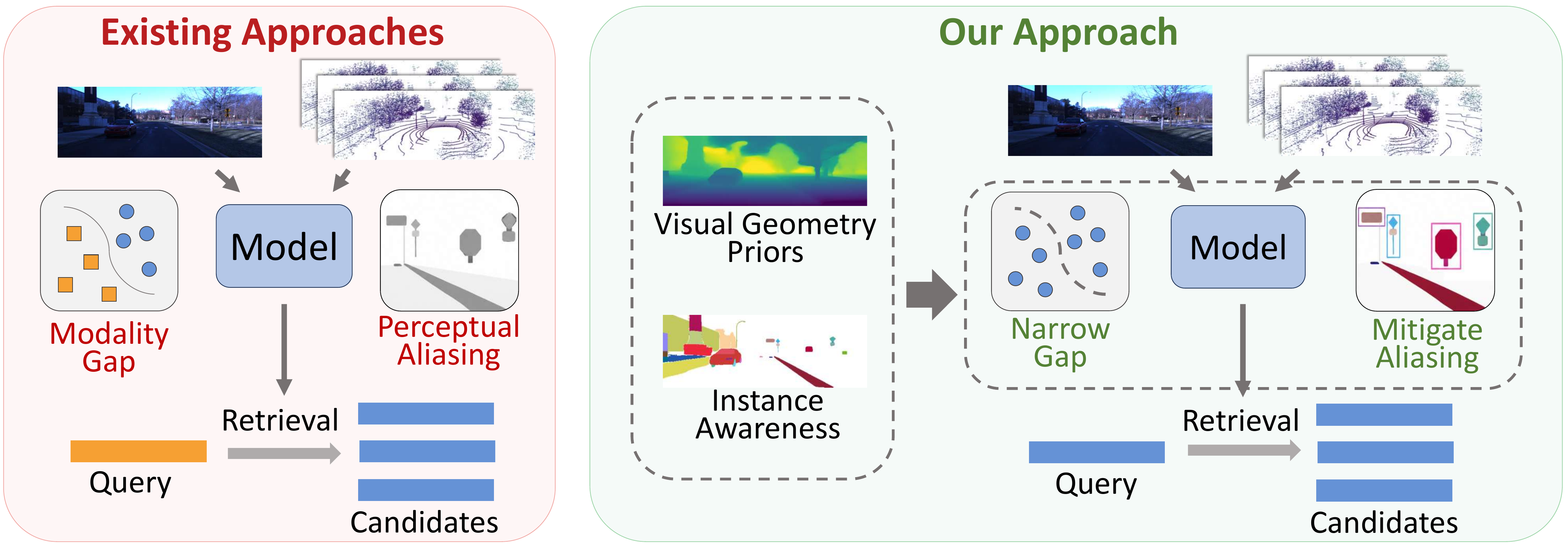}
  \vspace{-0.2cm}
  \caption{Existing CMPR approaches rely on global descriptors and are vulnerable to modality gaps and perceptual aliasing. Our proposed G2IA mitigates these issues using a retrieval-and-verification process with visual geometry priors and instance awareness.}
  \label{fig:motivation}
  \vspace{-0.3cm}
\end{figure}

Motivated by this insight, as illustrated in Fig.~\ref{fig:motivation}, we propose a geometry-guided instance-aware framework dubbed G2IA for image-to-point-cloud place recognition. Firstly, G2IA learns to adapt visual geometry priors from Visual Geometry Grounded Transformer (VGGT)~\cite{wang2025vggt} and instance features to construct place descriptors, making image queries more compatible with LiDAR-derived map representations. Secondly, we re-rank the retrieved candidates by explicitly checking whether local instance shapes and their relative spatial layouts are consistent across modalities. This design preserves the efficiency of global retrieval while introducing fine-grained verification for perceptually ambiguous urban scenes. The main contributions of our work are threefold:

\begin{itemize}

\item We propose G2IA, a geometry-guided instance-aware framework that couples efficient cross-modal retrieval with fine-grained candidate refinement for image-to-point-cloud place recognition. Extensive experiments on public datasets demonstrate its state-of-the-art performance against the existing CMPR baselines.

\item We incorporate VGGT-derived visual geometry priors and instance masks to capture scene-level cues and object-level structures for place descriptor generation, reducing the modality gap between RGB queries and LiDAR representations.

\item We refine retrieved results by jointly verifying the consistency of local instance shapes and relative spatial layouts between query and database candidates, improving robustness in perceptually similar scenes where global descriptors are ambiguous.

\end{itemize}

\section{Related Work}

\textbf{Image-to-Point-Cloud Place Recognition.} Cross-modal place recognition localizes a query from one sensor modality in a reference map built from another~\cite{kang2025opal,xia2024uniloc, xia2024text2loc}. In the image-to-point-cloud setting, the query is an RGB image and the database contains LiDAR point clouds. Prior work has reduced the modality gap by converting one modality into another or by learning modality-invariant descriptors. 
I2P-Rec~\cite{I2p-rec} is an early representative of the conversion-based paradigm, which estimates image depth to reconstruct 3D points and then performs retrieval in the BEV space. CrossBEV-PR~\cite{xu2025crossbev} also attends to BEV representation, but further transfers BEV priors from point clouds to visual networks through feature distillation. (LC)$^2$~\cite{lc2} instead utilizes a depth estimator to generate unscaled depth, which is encoded along with LiDAR range images to produce localization descriptors. To better align vision perception and LiDAR observations, ModaLink~\cite{xie2024modalink} implements a FoV overlap crop on the depth images, which inspires the data preprocessing of the following InsCMPR~\cite{jiao2025inscmpr}.
These methods mainly use modality conversion or depth/BEV priors for descriptor-level alignment. In contrast, G2IA leverages visual geometry priors and instance features to enhance the cross-modal alignment in both retrieval and refinement. 

\textbf{Foundation Models for Cross-Modal Place Recognition.} Foundation models have been increasingly explored in place recognition across visual, LiDAR, and multimodal settings~\cite{wang2022salient,keetha2023anyloc,lyu2024tell, jungimlpr, xu2026vggt}. For cross-modal place recognition, 
LIP-Loc~\cite{liploc} adopts a CLIP-based~\cite{radford2021learning} framework for LiDAR-image alignment, and CMVM~\cite{yao2025monocular} introduces a cross-modal state-space model with multi-view matching based on VMamba~\cite{liu2024vmamba}. InsCMPR~\cite{jiao2025inscmpr} further explores cross-modal retrieval by deploying MobileSAM~\cite{zhang2023faster} on images and depth maps for mask encoding.
While prior methods mainly adopt foundation models for generalizable vision feature extraction, geometry-grounded vision models with scene-level layout awareness remain underexplored in CMPR. Therefore, in this work, G2IA incorporates the geometry priors and instance features from two trendy foundation models, VGGT~\cite{wang2025vggt} and Objects~\cite{chen2025sam}, to support initial retrieval and candidate refinement in cross-modal place recognition.

\section{Method}

\begin{figure*}[t]
  \centering
  \includegraphics[width=0.95\linewidth]{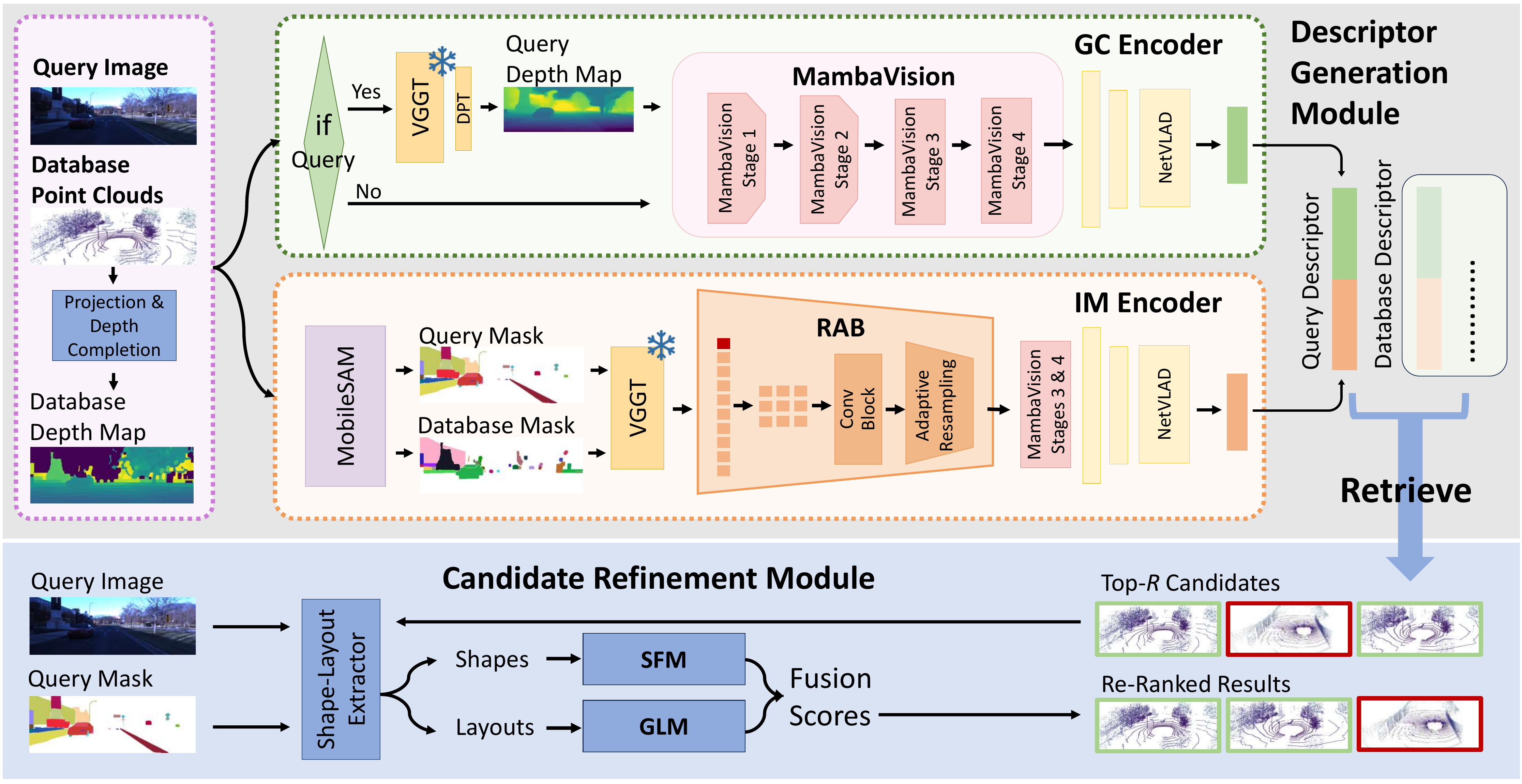}
  \vspace{-0.2cm}
  \caption{Overview of G2IA. The descriptor generation module maps RGB images and LiDAR point clouds into a shared descriptor space for efficient retrieval. The candidate refinement module then re-ranks the top-$R$ point-cloud candidates by final fusion scores.}
  \label{fig:pipeline}
  \vspace{-0.3cm}
\end{figure*}

\subsection{Overview}

To mitigate the modality gap and perceptual aliasing discussed in Sec.~\ref{sec:intro}, our G2IA formulates image-to-point-cloud place recognition as a retrieval-and-verification process. As shown in Fig.~\ref{fig:pipeline}, the framework consists of two modules: the descriptor generation module (DGM) and the candidate refinement module (CRM). DGM maps RGB images and LiDAR point clouds into a shared descriptor space for efficient retrieval. 
After the top-$R$ candidates are retrieved by DGM, CRM re-ranks them by verifying whether local instance shapes and relative spatial layouts are consistent across modalities.
In this way, G2IA uses geometry priors and instance features to narrow the cross-modal representation gap in the retrieval stage and to suppress perceptual aliasing in the refinement stage.

\subsection{Descriptor Generation Module}

The DGM aims to produce global place descriptors, reducing the modality gap by representing RGB images and LiDAR point clouds through geometry priors and instance features. For database point clouds $P\in\mathbb{R}^{N_\mathrm{pc}\times 3}$, we project the points onto the image plane using known calibration.
The projection produces a sparse depth map, which is completed by a depth completion method~\cite{ku2018defense} to obtain a dense representation $D^P\in\mathbb{R}^{H\times W\times 1}$. 
The DGM contains two complementary branches. The geometric context (GC) encoder captures scene-level geometry-aware cues, while the instance mask (IM) encoder preserves object-level structures. Together, these branches retain the efficiency of global retrieval while preserving local cues that are useful for candidate refinement.

\paragraph{Geometric Context Encoder.}
For a query image $I\in\mathbb{R}^{H\times W\times 3}$, we adopt VGGT~\cite{wang2025vggt} with a DPT head~\cite{ranftl2021vision} to infer a virtual depth map $D^I\in\mathbb{R}^{H\times W\times 1}$ as geometry priors. Because monocular depth does not necessarily preserve metric scale, both $D^I$ and $D^P$ are normalized before being fed into downstream encoders.
The normalized depth maps are then encoded by a MambaVision backbone~\cite{hatamizadeh2025mambavision} initialized with ImageNet-pretrained weights.
The resulting features are projected by an MLP and aggregated by NetVLAD~\cite{arandjelovic2016netvlad} to produce a geometric context descriptor $G_\mathrm{GC}\in\mathbb{R}^{d}$.
By using VGGT-derived depth for the image inputs, the GC encoder explicitly incorporates visual geometry priors to guide cross-modal retrieval with aligned geometry-aware representation.

\paragraph{Instance Mask Encoder.}
The IM encoder complements geometric context with instance-level features that are less sensitive to repeated global appearances. 
We employ MobileSAM~\cite{zhang2023faster} on $I$ and $D^P$ to obtain foreground instance masks $\mathcal{M}^I$ and $\mathcal{M}^P$, respectively, filtering out large background regions such as sky and road. The masked visual signals are then processed by a frozen VGGT backbone to produce patch tokens. A resolution adaptation bridge (RAB) reshapes the token sequence into a spatial feature volume and aligns it with the subsequent MambaVision stages. The feature volume is then projected by an MLP and aggregated with NetVLAD to obtain an instance mask descriptor $G_\mathrm{IM}\in\mathbb{R}^{d}$. Ultimately, the two representations $G_\mathrm{GC}$ and $G_\mathrm{IM}$ are concatenated to form the final global descriptor $G\in\mathbb{R}^{2d}$. More architectural details are provided in the appendix.

We train DGM with lazy triplet loss, which is widely adopted in place recognition methods~\cite{uy2018pointnetvlad,xie2024modalink,jiao2025inscmpr}. For an anchor query descriptor $G_{\mathrm{query}}$, a positive LiDAR descriptor $G_{\mathrm{pos}}$, and $N_{\mathrm{neg}}$ negative LiDAR descriptors $\{G_{\mathrm{neg}}^{i}\}_{i=1}^{N_{\mathrm{neg}}}$, the loss is calculated by:
\begin{align}
\begin{split}
    \mathcal{L} & = [\beta+\left\|G_{\mathrm{query}}- G_{\mathrm{pos}}\right\|_2 \\
    & -\min_{1 \leq i \leq N_{\mathrm{neg}}}\left\|G_{\mathrm{query}}- G_{\mathrm{neg}}^{i}\right\|_2
    ]_{+},
\end{split}
\label{eq:loss}
\end{align}
where $[\cdot]_{+}$ denotes the hinge function, $\beta$ is the margin, and $\left\| \cdot \right\|_2$ computes Euclidean distance.
After training, LiDAR descriptors are indexed to construct the reference database $\mathcal{D}$. At inference time, the image descriptor retrieves the top-$R$ candidates $\mathcal{C}_R$ from $\mathcal{D}$ using FAISS~\cite{johnson2019billion}. These candidates are subsequently fed into CRM for instance-aware refinement.

\subsection{Candidate Refinement Module}

\begin{figure*}[t]
  \centering
  \includegraphics[width=0.80\linewidth]{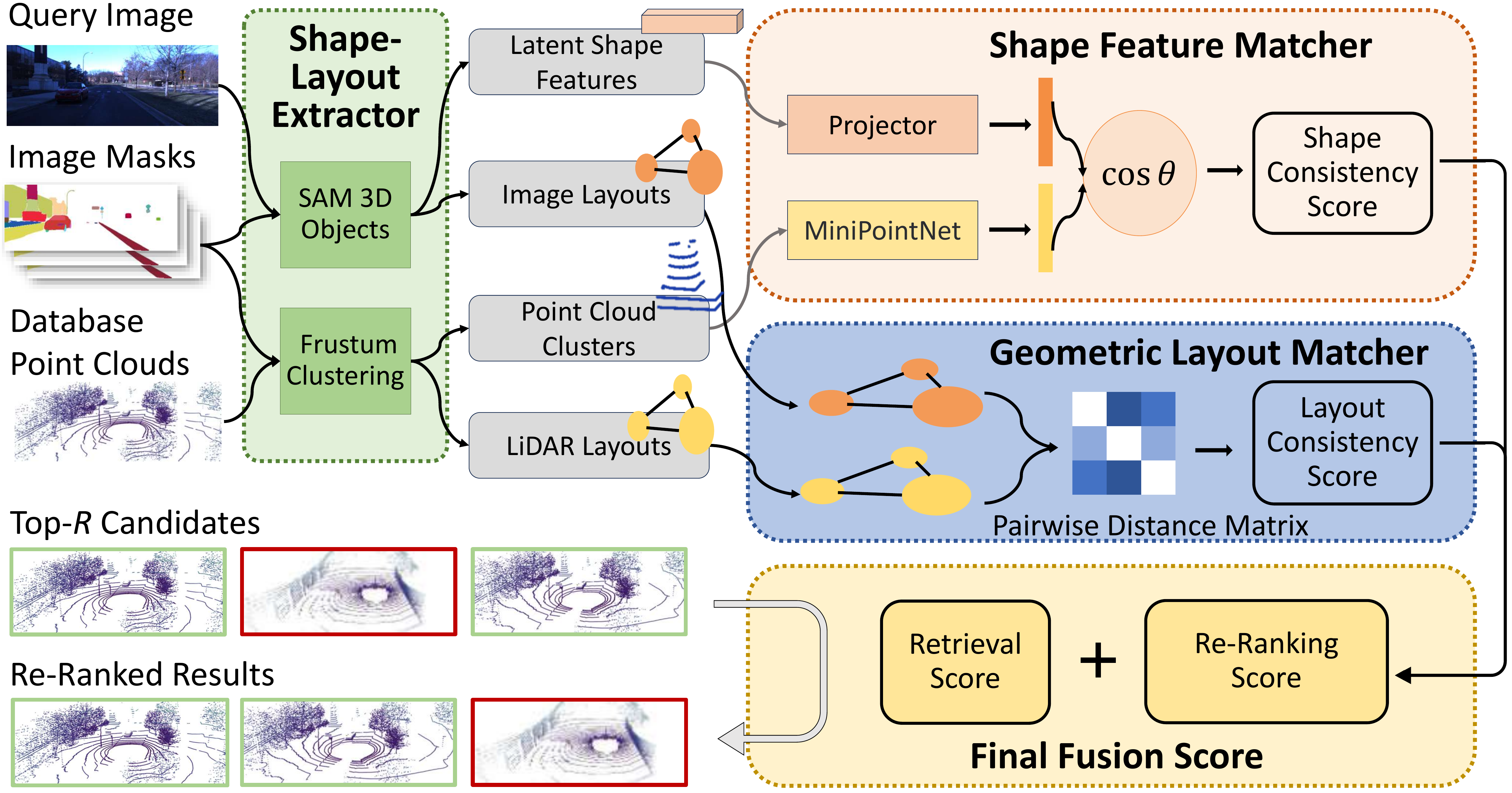}
  \vspace{-0.2cm}
  \caption{Candidate refinement module. Query instance masks guide the extraction of shape-layout cues and candidate point-cloud clusters. The shape feature matcher compares instance shape representations, and the geometric layout matcher compares relative arrangements. The re-ranking scores are fused with initial retrieval scores to produce the final refinement ranking.}
  \label{fig:IARM}
  \vspace{-0.3cm}
\end{figure*}

Although DGM improves cross-modal retrieval, global descriptors can still confuse different places with similar appearances. The CRM addresses this perceptual aliasing by re-ranking the top-$R$ candidates according to instance-level consistency. As detailed in Fig.~\ref{fig:IARM}, it checks whether the candidate point clouds preserve both the local instance shapes and the relative spatial layout observed in the query image.
The geometric layout matcher (GLM) assesses whether the relative spatial configuration of instances is consistent across modalities, while the shape feature matcher (SFM) measures whether the corresponding instance shapes are semantically compatible. 
The final fusion score combines the re-ranking score with the initial retrieval score, rejecting globally similar but locally inconsistent candidates.

\paragraph{Shape-Layout Extractor.}
To obtain image-side shape and layout cues, we adapt SAM 3D Objects~\cite{chen2025sam}, a pre-trained single-view object reconstruction model. Full 3D content generation is unnecessary for place recognition, so we use the intermediate inference pathway instead of final decoding subnetworks.
SAM 3D Objects is used to produce a latent shape feature $V_k$ for each instance, and an image-side layout $L^I$, which denotes the pairwise distance matrix of all instance centers within the query image $I$.

\begin{table*}[t]
\scriptsize
\centering
\captionsetup{aboveskip=2pt, belowskip=0pt}
\caption{Performance comparison on NCLT under different positive distance thresholds.}
\label{tab:nclt_comparison}
\setlength{\tabcolsep}{9pt}
\renewcommand{\arraystretch}{0.9}
\begin{tabular}{l|ccc|ccc|ccc|ccc}
\toprule
\multirow{2}{*}{Method}
& \multicolumn{3}{c|}{2012-02-05}
& \multicolumn{3}{c|}{2012-06-15}
& \multicolumn{3}{c|}{2013-02-23}
& \multicolumn{3}{c}{2013-04-05} \\
\cmidrule(lr){2-4}
\cmidrule(lr){5-7}
\cmidrule(lr){8-10}
\cmidrule(lr){11-13}
&
AR@1 & AR@5 & AR@10
& AR@1 & AR@5 & AR@10
& AR@1 & AR@5 & AR@10
& AR@1 & AR@5 & AR@10 \\
\midrule
\multicolumn{13}{c}{\textbf{Threshold = 0.5m}} \\
\midrule
LIP-Loc
& 51.90 & 77.88 & 86.49
& 19.01 & 38.83 & 49.52
& 19.33 & 36.13 & 45.35
& 32.17 & 55.27 & 66.82 \\
ModaLink
& 71.05 & 94.32 & 96.57
& 25.99 & 42.12 & 47.20
& 54.46 & 78.04 & 83.90
& 67.67 & 89.60 & 93.51 \\
InsCMPR
& \underline{73.53} & \underline{94.37} & \underline{97.41}
& \underline{52.50} & \underline{75.42} & \underline{81.41}
& \underline{55.46} & \underline{80.97} & \underline{88.51}
& \underline{68.81} & \underline{91.25} & \textbf{95.48} \\
\textbf{G2IA}
& \textbf{76.17} & \textbf{95.37} & \textbf{97.68}
& \textbf{75.81} & \textbf{94.65} & \textbf{97.05}
& \textbf{73.47} & \textbf{91.57} & \textbf{94.60}
& \textbf{74.32} & \textbf{91.94} & \underline{94.67} \\
\midrule
\multicolumn{13}{c}{\textbf{Threshold = 10m}} \\
\midrule
LIP-Loc
& 79.63 & 91.96 & 95.18
& 47.91 & 65.58 & 73.19
& 39.63 & 58.67 & 67.71
& 59.93 & 78.24 & 85.27 \\
ModaLink
& 91.30 & 97.70 & 98.80
& 42.37 & 56.97 & 64.72
& 72.43 & 87.17 & 90.70
& 85.80 & 95.60 & 97.30 \\
InsCMPR
& \underline{91.53} & \underline{97.91} & \underline{99.09}
& \underline{69.52} & \underline{83.37} & \underline{88.17}
& \underline{74.81} & \underline{91.23} & \underline{95.67}
& \underline{87.20} & \underline{96.51} & \textbf{98.21} \\
\textbf{G2IA}
& \textbf{96.64} & \textbf{98.68} & \textbf{99.31}
& \textbf{96.22} & \textbf{98.59} & \textbf{99.14}
& \textbf{93.76} & \textbf{97.03} & \textbf{97.79}
& \textbf{93.36} & \textbf{96.95} & \underline{97.64} \\
\bottomrule
\end{tabular}
\begin{flushleft}
\scriptsize
\vspace{-0.1cm}
\,\,\,\,\,\,The best and secondary results are highlighted in \textbf{bold black} and \underline{underline}, respectively.
\end{flushleft}
\vspace{-0.3cm}
\end{table*}

For each retrieved point-cloud candidate $C_j\in\mathcal{C}_R$, we posit a \textit{same place} hypothesis: $C_j$ is treated as being collected in the same place of the query frame. Consequently, the cloud points of $C_j$ are projected onto the query image plane using the camera intrinsics. For each mask $m_k^I\in \mathcal{M}^I$, only points falling inside the corresponding image frustum are retained. For point-cloud-side shape and layout, DBSCAN~\cite{ester1996density} is then applied to remove background outliers and isolate the dominant point cluster, yielding point cluster $Q_{j,k}$ and a point-cloud-side layout $L^P_{j}$. Incorrect candidates tend to produce fragmented or misaligned clusters, while correct candidates produce coherent clusters that preserve the query's instance structure.

\paragraph{Geometric Layout Matcher.}
The GLM measures relative spatial consistency. To reduce the effect of scale ambiguity in single-view geometry, pairwise Euclidean distances in $L^I$ and $L^P_{j}$ are normalized within each modality. Let $\bar{e}^{I}_{a,b}$ and $\bar{e}^{P}_{a,b}$ denote the normalized pairwise distances between instances $a$ and $b$ of image-side and point-cloud-side layouts, respectively. The layout consistency score across all $K$ instances is:
\begin{align}
S_{\mathrm{LC}} = \frac{2}{K(K-1)} \sum_{a=1}^{K-1} \sum_{b=a+1}^{K} \exp \left( -\sigma \left| \bar{e}^{I}_{a,b} - \bar{e}^{P}_{a,b} \right| \right),
\label{eq:score_glm}
\end{align}
where $\sigma$ controls the penalty for structural discrepancy.

\paragraph{Shape Feature Matcher.}
The SFM measures shape consistency. The image-side latent shape feature $V_k$ is mapped by $\phi(\cdot)$, which consists of spatial pooling and non-linear projections. The point cluster $Q_{j,k}$ is encoded by MiniPointNet~\cite{qi2017pointnet}, denoted as $\psi(\cdot)$. 
The shape consistency score is computed by:
\begin{align}
S_{\mathrm{SC}} =
\frac{1}{K}
\sum_{k=1}^{K}
\left\langle
\frac{\phi(V_k)}
{\left\| \phi(V_k) \right\|_2},
\frac{\psi(Q_{j,k})}
{\left\| \psi(Q_{j,k}) \right\|_2}
\right\rangle,
\label{eq:score_sfm}
\end{align}
where $\langle \cdot,\cdot \rangle$ denotes the inner product operation. The re-ranking score is composed of the layout consistency score and the shape consistency score.
Let $d_j$ be the global descriptor distance between the query and $C_j$. We convert it into an initial retrieval score:
\begin{align}
S_{\mathrm{IR}} = 1 - \frac{d_j}{\max_{j \in \mathcal{C}_R} d_j + \epsilon},
\label{eq:score_base}
\end{align}
where $\epsilon$ is a small numerical constant. Finally, we combine $S_{\mathrm{IR}}$ with the re-ranking score to form the final fusion score:
\begin{align}
S_{\mathrm{Fusion}} = w_{\mathrm{IR}} S_{\mathrm{IR}} + w_{\mathrm{LC}} S_{\mathrm{LC}} + w_{\mathrm{SC}} S_{\mathrm{SC}},
\label{eq:score_fusion}
\end{align}
where $w_{\mathrm{IR}}$, $w_{\mathrm{LC}}$, and $w_{\mathrm{SC}}$ are score weights. Database candidates $\mathcal{C}_R$ are sorted by $S_{\mathrm{Fusion}}$ to obtain the final place recognition results.

\begin{table*}[t]
\scriptsize
\centering
\captionsetup{aboveskip=2pt, belowskip=0pt}
\renewcommand\arraystretch{0.9}
\caption{Performance comparison on KITTI under different positive distance thresholds.}
\label{tab:kitti_comparison}
\setlength{\tabcolsep}{5.2pt}
\begin{tabular}{l|ccc|ccc|ccc|ccc|ccc}
\toprule
\multirow{2}{*}{Method}
& \multicolumn{3}{c|}{00}
& \multicolumn{3}{c|}{02}
& \multicolumn{3}{c|}{05}
& \multicolumn{3}{c|}{06}
& \multicolumn{3}{c}{08} \\
\cmidrule(lr){2-4}
\cmidrule(lr){5-7}
\cmidrule(lr){8-10}
\cmidrule(lr){11-13}
\cmidrule(lr){14-16}
&
AR@1 & AR@5 & AR@10
& AR@1 & AR@5 & AR@10
& AR@1 & AR@5 & AR@10
& AR@1 & AR@5 & AR@10
& AR@1 & AR@5 & AR@10 \\
\midrule
\multicolumn{16}{c}{\textbf{Threshold = 0.5m}} \\
\midrule
LIP-Loc
& 25.18 & 44.71 & 53.08
& 0.56 & 1.93 & 3.58
& 3.95 & 10.43 & 16.66
& 14.17 & 33.33 & 48.96
& 3.37 & 7.89 & 11.37 \\
ModaLink
& 81.78 & 92.17 & 94.74
& 40.34 & 58.93 & 64.72
& 53.84 & 76.10 & 81.73
& 80.93 & 95.10 & 96.02
& 58.97 & 79.44 & 84.74 \\
InsCMPR
& \underline{89.73} & \underline{96.81} & \underline{98.76}
& \underline{63.70} & \underline{80.41} & \underline{85.16}
& \underline{72.15} & \underline{89.46} & \underline{93.12}
& \underline{84.56} & \underline{95.07} & \textbf{98.00}
& \underline{80.59} & \underline{92.29} & \underline{94.77} \\
\textbf{G2IA}
& \textbf{97.59} & \textbf{99.22} & \textbf{99.22}
& \textbf{85.07} & \textbf{88.94} & \textbf{89.52}
& \textbf{87.98} & \textbf{92.76} & \textbf{93.30}
& \textbf{89.65} & \textbf{95.19} & \underline{96.37}
& \textbf{94.65} & \textbf{97.74} & \textbf{97.89} \\
\midrule
\multicolumn{16}{c}{\textbf{Threshold = 10m}} \\
\midrule
LIP-Loc
& 49.90 & 67.88 & 76.05
& 4.68 & 12.36 & 18.47
& 24.48 & 46.69 & 58.60
& 62.40 & 94.73 & 98.82
& 15.11 & 27.95 & 36.82 \\
ModaLink
& 90.99 & 97.30 & 98.77
& 51.09 & 71.63 & 80.12
& 73.77 & 87.53 & 91.42
& 93.28 & 99.64 & \textbf{99.91}
& 71.75 & 88.74 & 92.73 \\
InsCMPR
& \underline{95.84} & \underline{99.02} & \underline{99.80}
& \underline{80.17} & \underline{90.24} & \underline{93.39}
& \underline{91.45} & \underline{95.87} & \underline{97.01}
& \underline{95.10} & \textbf{99.67} & \textbf{99.91}
& \underline{91.38} & \underline{96.00} & \underline{97.49} \\
\textbf{G2IA}
& \textbf{99.22} & \textbf{99.87} & \textbf{99.87}
& \textbf{89.07} & \textbf{92.31} & \textbf{93.44}
& \textbf{94.02} & \textbf{96.34} & \textbf{97.10}
& \textbf{96.19} & \underline{98.37} & \underline{98.82}
& \textbf{97.45} & \textbf{98.97} & \textbf{99.34} \\
\bottomrule
\end{tabular}
\vspace{-0.3cm}
\end{table*}

\section{Experiments}
\subsection{Experimental Setups}

\begin{table*}[t]
\scriptsize
\centering
\captionsetup{aboveskip=2pt, belowskip=0pt}
\caption{Zero-shot cross-dataset evaluation on KITTI.}
\label{tab:kitti_zeroshot}
\setlength{\tabcolsep}{5.2pt}
\renewcommand{\arraystretch}{0.9}
\begin{tabular}{l|ccc|ccc|ccc|ccc|ccc}
\toprule
\multirow{2}{*}{Method}
& \multicolumn{3}{c|}{00}
& \multicolumn{3}{c|}{02}
& \multicolumn{3}{c|}{05}
& \multicolumn{3}{c|}{06}
& \multicolumn{3}{c}{08} \\
\cmidrule(lr){2-4}
\cmidrule(lr){5-7}
\cmidrule(lr){8-10}
\cmidrule(lr){11-13}
\cmidrule(lr){14-16}
&
AR@1 & AR@5 & AR@10
& AR@1 & AR@5 & AR@10
& AR@1 & AR@5 & AR@10
& AR@1 & AR@5 & AR@10
& AR@1 & AR@5 & AR@10 \\
\midrule
\multicolumn{16}{c}{\textbf{Threshold = 0.5m}} \\
\midrule
LIP-Loc
& \underline{25.18} & \underline{44.71} & \underline{53.08}
& 0.56 & 1.93 & 3.58
& 3.95 & 10.43 & 16.66
& \underline{14.17} & \underline{33.33} & \underline{48.96}
& 3.37 & 7.89 & 11.37 \\
ModaLink
& 0.65 & 1.43 & 2.99
& 0.19 & 0.69 & 0.92
& 0.22 & 0.91 & 1.85
& 5.27 & 12.17 & 17.26
& 0.20 & 0.52 & 0.84 \\
InsCMPR
& 7.93 & 16.84 & 22.82
& \underline{3.37} & \underline{7.97} & \underline{10.96}
& \underline{7.57} & \underline{13.76} & \underline{17.13}
& 6.09 & 14.71 & 20.71
& \underline{4.45} & \underline{9.48} & \underline{13.34} \\
\textbf{G2IA}
& \textbf{69.96} & \textbf{80.10} & \textbf{81.40}
& \textbf{59.38} & \textbf{70.01} & \textbf{71.58}
& \textbf{59.29} & \textbf{69.58} & \textbf{72.91}
& \textbf{79.38} & \textbf{88.10} & \textbf{89.19}
& \textbf{62.79} & \textbf{71.46} & \textbf{72.49} \\
\midrule
\multicolumn{16}{c}{\textbf{Threshold = 10m}} \\
\midrule
LIP-Loc
& \underline{49.90} & \underline{67.88} & \underline{76.05}
& 4.68 & 12.36 & 18.47
& \underline{24.48} & \underline{46.69} & \underline{58.60}
& \underline{62.40} & \underline{94.73} & \underline{98.82}
& \underline{15.11} & \underline{27.95} & \underline{36.82} \\
ModaLink
& 13.76 & 17.33 & 21.61
& 3.35 & 6.05 & 8.32
& 6.27 & 10.47 & 15.47
& 54.13 & 79.38 & 98.73
& 6.98 & 12.36 & 16.09 \\
InsCMPR
& 18.27 & 30.82 & 39.53
& \underline{8.44} & \underline{16.24} & \underline{21.12}
& 16.99 & 26.98 & 32.42
& 15.89 & 25.89 & 35.15
& 9.36 & 17.05 & 23.07 \\
\textbf{G2IA}
& \textbf{78.09} & \textbf{87.06} & \textbf{89.86}
& \textbf{68.01} & \textbf{78.88} & \textbf{82.58}
& \textbf{74.28} & \textbf{84.93} & \textbf{88.88}
& \textbf{84.56} & \textbf{92.64} & \textbf{93.73}
& \textbf{71.58} & \textbf{80.25} & \textbf{83.57} \\
\bottomrule
\end{tabular}
\vspace{-0.3cm}
\end{table*}

\begin{figure*}[t]
  \centering
  \includegraphics[width=0.80\linewidth]{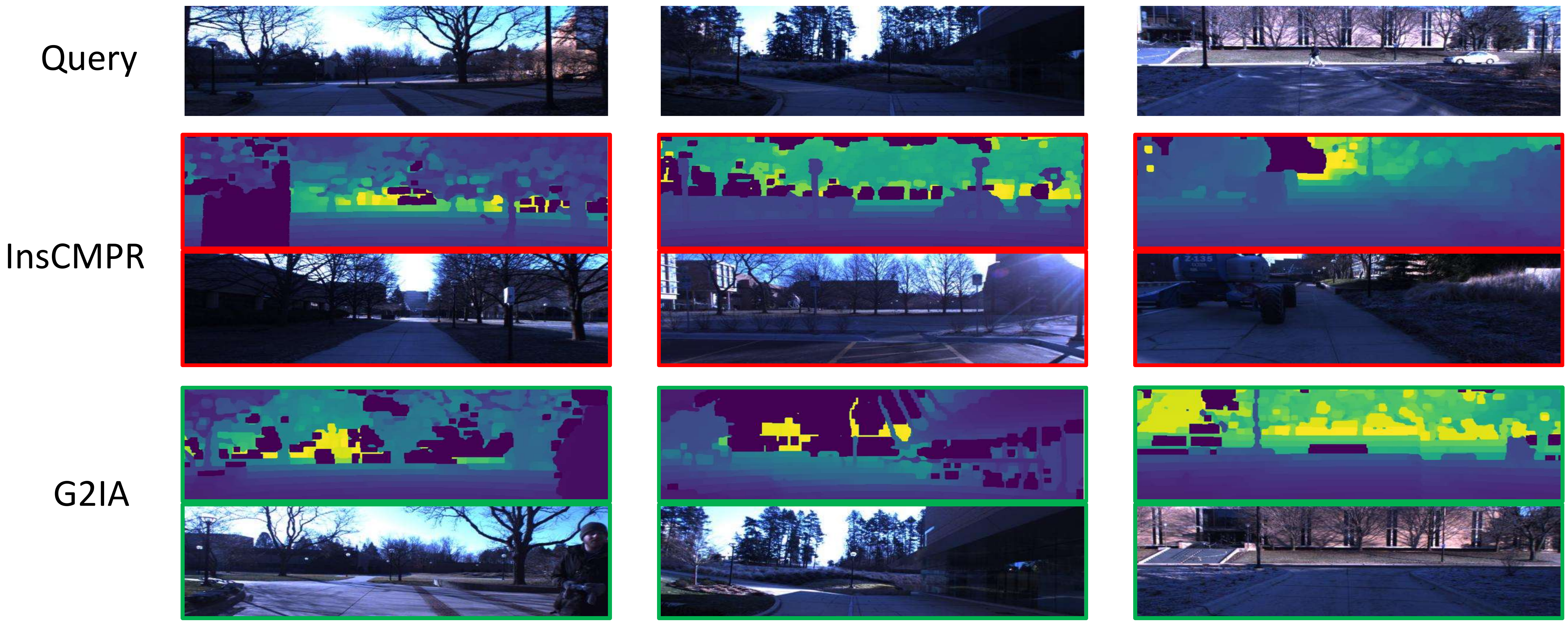}
  \vspace{-0.2cm}
  \caption{Visualization of the results from our method and InsCMPR. Green boxes denote correct top-$1$ retrieved places, while red boxes denote incorrect ones. For clearer comparison, we show RGB images and completed point clouds for each frame.}
  \label{fig:compare_with_inscmpr}
  \vspace{-0.3cm}
\end{figure*}

\paragraph{Datasets and Baselines.}
We evaluate G2IA on two public outdoor autonomous driving datasets, NCLT~\cite{carlevaris2016university} and KITTI~\cite{geiger2013vision}, following recent image-to-point-cloud place recognition protocols~\cite{jiao2025inscmpr, xie2024modalink}. For NCLT, sequence ``2012-01-08'' is used for training, and sequences ``2012-02-05'', ``2012-06-15'', ``2013-02-23'', and ``2013-04-05'' are used for testing. For KITTI, the first 3000 frames of sequence 00 are used for training. The remaining frames of sequence 00, and sequences 02, 05, 06, and 08 are all used for evaluation.
Besides, we compare G2IA against representative baselines including LIP-Loc~\cite{liploc}, ModaLink~\cite{xie2024modalink}, and InsCMPR~\cite{jiao2025inscmpr}. For fairness, all methods use forward-facing image queries and the settings from their original papers.
The training details of all the baselines and our G2IA are provided in the appendix.

\paragraph{Inference and Evaluation.}
During inference, each query image first retrieves the top-$20$ ($R=20$) point-cloud candidates. The CRM then re-ranks these candidates according to the final fused score, where we set $\sigma$ = 5, $\epsilon=1e-6$, $w_{\mathrm{IR}} = 0.2$, $w_{\mathrm{LC}} = 0.2$, and $w_{\mathrm{SC}} = 0.6$. The selection of these weights is ablated in the appendix.
To quantify retrieval performance, we report average recall (AR@1/5/10) under both 0.5\,m and 10\,m positive distance thresholds, corresponding to strict localization accuracy and more tolerant retrieval, respectively.

\begin{table*}[t]
\scriptsize
\centering
\captionsetup{aboveskip=2pt, belowskip=0pt}
\caption{Ablation study of VGGT integration on KITTI under the 10\,m positive distance threshold.}
\label{tab:kitti_ablation_vggt_10m}
\setlength{\tabcolsep}{3pt}
\renewcommand{\arraystretch}{0.9}
\begin{tabular}{cc|ccc|ccc|ccc|ccc|ccc}
\toprule
\multirow{2}{*}{\begin{tabular}{c}VGGT in\\GC Encoder\end{tabular}}
& \multirow{2}{*}{\begin{tabular}{c}VGGT in\\IM Encoder\end{tabular}}
& \multicolumn{3}{c|}{00}
& \multicolumn{3}{c|}{02}
& \multicolumn{3}{c|}{05}
& \multicolumn{3}{c|}{06}
& \multicolumn{3}{c}{08} \\
\cmidrule(lr){3-5}
\cmidrule(lr){6-8}
\cmidrule(lr){9-11}
\cmidrule(lr){12-14}
\cmidrule(lr){15-17}
&
& AR@1 & AR@5 & AR@10
& AR@1 & AR@5 & AR@10
& AR@1 & AR@5 & AR@10
& AR@1 & AR@5 & AR@10
& AR@1 & AR@5 & AR@10 \\
\midrule
$\times$ & $\times$
& 92.91 & 95.58 & 96.49
& 73.25 & 82.00 & 84.34
& 85.55 & 91.78 & 93.48
& 93.46 & 97.73 & 98.00
& 92.16 & 96.76 & 97.69 \\
$\times$ & $\checkmark$
& 96.55 & 98.76 & 99.41
& 78.26 & 87.22 & 89.04
& 90.04 & 94.86 & 96.02
& 93.73 & 98.09 & 98.73
& 94.79 & 98.11 & 98.55 \\
$\checkmark$ & $\times$
& 98.11 & 99.22 & 99.54
& 86.92 & 92.02 & 93.16
& 93.63 & 96.29 & 97.04
& 94.82 & 98.09 & 98.73
& 97.10 & 98.62 & 99.31 \\
$\checkmark$ & $\checkmark$
& \textbf{99.22} & \textbf{99.87} & \textbf{99.87}
& \textbf{89.07} & \textbf{92.31} & \textbf{93.44}
& \textbf{94.02} & \textbf{96.34} & \textbf{97.10}
& \textbf{96.19} & \textbf{98.37} & \textbf{98.82}
& \textbf{97.45} & \textbf{98.97} & \textbf{99.34} \\
\bottomrule
\end{tabular}
\vspace{-0.3cm}
\end{table*}

\begin{table*}[t]
\scriptsize
\setlength{\tabcolsep}{4.5pt}
\centering
\captionsetup{aboveskip=2pt, belowskip=0pt}
\renewcommand{\arraystretch}{0.9}
\caption{Ablation study of the candidate refinement module on KITTI under the 10\,m positive distance threshold.}
\label{tab:kitti_ablation_re-rank_10m}
\begin{tabular}{cc|ccc|ccc|ccc|ccc|ccc}
\toprule
\multirow{2}{*}{GLM}
& \multirow{2}{*}{SFM}
& \multicolumn{3}{c|}{00}
& \multicolumn{3}{c|}{02}
& \multicolumn{3}{c|}{05}
& \multicolumn{3}{c|}{06}
& \multicolumn{3}{c}{08} \\
\cmidrule(lr){3-5}
\cmidrule(lr){6-8}
\cmidrule(lr){9-11}
\cmidrule(lr){12-14}
\cmidrule(lr){15-17}
&
& AR@1 & AR@5 & AR@10
& AR@1 & AR@5 & AR@10
& AR@1 & AR@5 & AR@10
& AR@1 & AR@5 & AR@10
& AR@1 & AR@5 & AR@10 \\
\midrule
$\times$ & $\times$
& 95.32 & 99.22 & 99.74
& 83.37 & 90.20 & 92.29
& 86.78 & 94.75 & 96.52
& 88.74 & 96.91 & 98.37
& 90.86 & 97.57 & 98.75 \\
$\times$ & $\checkmark$
& 98.30 & 99.80 & 99.87
& 86.65 & 91.86 & 93.13
& 92.00 & 95.69 & 96.85
& 92.73 & 97.55 & 98.64
& 94.15 & 98.40 & 99.09 \\
$\checkmark$ & $\times$
& 98.83 & 99.87 & 99.87
& 87.84 & 92.14 & 93.21
& 92.29 & 96.05 & 97.03
& 95.19 & 98.09 & 98.64
& 96.51 & 98.82 & 99.24 \\
$\checkmark$ & $\checkmark$
& \textbf{99.22} & \textbf{99.87} & \textbf{99.87}
& \textbf{89.07} & \textbf{92.31} & \textbf{93.44}
& \textbf{94.02} & \textbf{96.34} & \textbf{97.10}
& \textbf{96.19} & \textbf{98.37} & \textbf{98.82}
& \textbf{97.45} & \textbf{98.97} & \textbf{99.34} \\
\bottomrule
\end{tabular}
\vspace{-0.3cm}
\end{table*}

\begin{figure*}[t]
  \centering
  \includegraphics[width=0.80\linewidth]{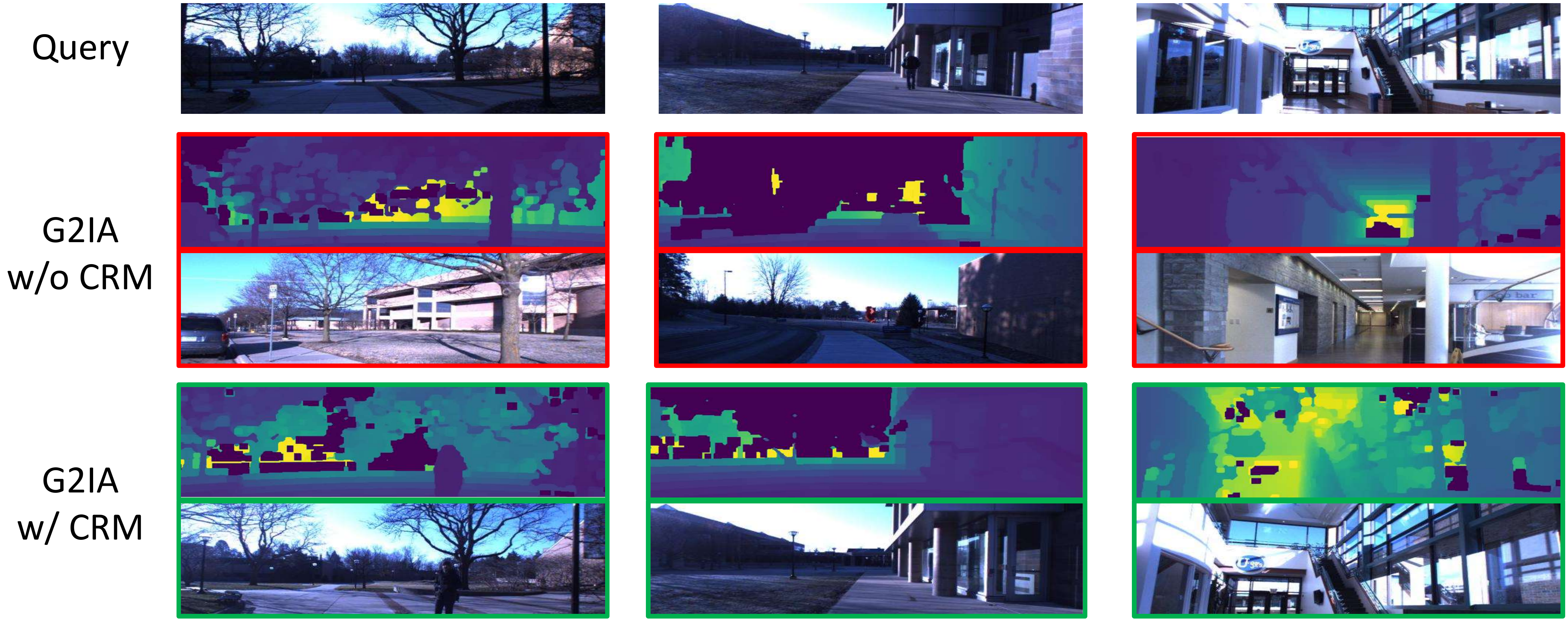}
  \vspace{-0.2cm}
  \caption{Qualitative effectiveness of the candidate refinement.}
  \label{fig:compare_with_rerank}
  \vspace{-0.3cm}
\end{figure*}

\subsection{Performance Comparison}

\paragraph{Comparison on the NCLT Dataset.}
We first compare G2IA with state-of-the-art baselines on NCLT. As shown in Tab.~\ref{tab:nclt_comparison}, G2IA significantly outperforms the baselines across all sequences and both thresholds. Under the commonly used 10\,m threshold, it improves AR@1 over the strongest baseline InsCMPR by 5.11\%,  26.70\%, 18.95\%, 6.16\% on sequences ``2012-02-05'', ``2012-06-15'', ``2013-02-23'', and ``2013-04-05'', respectively.
Under the stricter 0.5\,m threshold, the gains remain clear at 2.64\%, 23.31\%, 18.01\%, and 5.51\%. 
Fig.~\ref{fig:compare_with_inscmpr} qualitatively compares G2IA with InsCMPR in challenging scenes with similar roads, facades, intersections, and object arrangements. The baseline often retrieves visually similar but incorrect places, while G2IA achieves more correct top-$1$ matches. More visualization results are provided in the appendix.

\paragraph{Comparison on the KITTI Dataset.}
In Tab.~\ref{tab:kitti_comparison}, G2IA also achieves the best AR@1 on all five evaluation sequences of KITTI, reaching AR@1 of 99.22\%, 89.07\%, 94.02\%, 96.19\%, and 97.45\% under the 10\,m threshold, and 97.59\%, 85.07\%, 87.98\%, 89.65\%, and 94.65\% under the 0.5\,m threshold. These results show that G2IA improves both coarse and fine-grained retrieval, further indicating that modality alignment and candidate verification are necessary for reliable CMPR.
We also evaluate cross-dataset generalization by training models on NCLT and directly testing them on KITTI without fine-tuning. As shown in Tab.~\ref{tab:kitti_zeroshot}, G2IA achieves the highest recognition accuracy across all sequences. The experimental details are provided in the appendix.

\subsection{Ablation Studies}

\paragraph{Ablation on VGGT in DGM.}
We conduct ablation studies to evaluate how VGGT contributes to the GC and IM encoders of DGM. As shown in Tab.~\ref{tab:kitti_ablation_vggt_10m}, removing VGGT from both encoders leads to a clear performance drop, confirming that visual geometry priors are important for aligning image and LiDAR depth representations. Introducing VGGT into the GC encoder brings substantial gains, showing that geometry-aware global context helps reduce the modality gap. Using VGGT in the IM encoder also improves performance by encoding instance-level cues. The best results are obtained when VGGT is used in both encoders, indicating that geometry-aware global context and instance representations are complementary. 

\paragraph{Ablation on CRM.}
As shown in Tab.~\ref{tab:kitti_ablation_re-rank_10m}, we assess the contributions of GLM and SFM within CRM. 
When a matcher is disabled, its corresponding score weight is set to zero while the remaining terms are kept unchanged. The results show that both GLM and SFM contribute to performance, and combining both matchers consistently achieves the best results. This demonstrates that integrating spatial layout and instance shapes is critical for improving cross-modal place recognition.
Fig.~\ref{fig:compare_with_rerank} further visualizes the effect of the CRM. Without re-ranking, G2IA may still retrieve candidates that are globally similar to the query but locally inconsistent. After applying re-ranking, the model corrects these failure cases by performing finer instance-level verification. 
Please refer to the appendix for additional ablation studies on more datasets and distance thresholds.

\section{Conclusion}
\label{sec:conclusion}

We propose G2IA, a geometry-guided instance-aware framework for image-to-point-cloud place recognition. The descriptor generation module improves cross-modal retrieval by using visual geometry priors and instance features to reduce the modality gap between RGB queries and LiDAR point-cloud maps. The candidate refinement module then verifies local shape consistency and spatial layout consistency across modalities, improving robustness in urban scenes with repeated appearance and ambiguous global structure. Experiments on NCLT and KITTI demonstrate the effectiveness and generalization ability of G2IA, while ablation studies confirm the complementary roles of semantic feature matching and geometric layout matching.

\section{Limitations}
Although G2IA improves image-to-point-cloud place recognition, several limitations remain. Its image-side layout is derived from single-view geometry. While GLM uses normalized pairwise distances, such normalization may discard useful metric cues for visual place description. In addition, the framework also assumes accurate camera-LiDAR calibration. Inaccurate calibration results may lead to performance degradation. Therefore, in future work, we plan to integrate powerful metric depth estimation methods into G2IA and assess its robustness to calibration noise.

\bibliographystyle{ieeetr}


\footnotesize{
\bibliography{new}}

\end{document}

%% file: stachnisslab-latex.tex
\usepackage{graphics}           
\usepackage{times}              
\usepackage{amsmath}            
\usepackage{amssymb}            
\usepackage{graphicx}
\usepackage{algorithm}
\usepackage[noend]{algpseudocode}
\usepackage{booktabs}
\usepackage{color}
\definecolor{instructioncolor}{rgb}{.5,.5,.5}

\usepackage[font=small]{caption}

\def\eqref#1{Eq.~(\ref{#1})}

\makeatletter
\usepackage{xspace}
\DeclareRobustCommand\onedot{\futurelet\@let@token\@onedot}
\def\@onedot{\ifx\@let@token.\else.\null\fi\xspace}

\makeatother

\usepackage{array}
\newcolumntype{L}[1]{>{\raggedright\let\newline\\\arraybackslash\hspace{0pt}}m{#1}}
\newcolumntype{C}[1]{>{\centering\let\newline\\\arraybackslash\hspace{0pt}}m{#1}}
\newcolumntype{R}[1]{>{\raggedleft\let\newline\\\arraybackslash\hspace{0pt}}m{#1}}